%% file: sdt.tex
\documentclass[sigconf]{acmart}
\usepackage{multirow}
\usepackage{balance} 
\AtBeginDocument{%
  \providecommand\BibTeX{{%
    \normalfont B\kern-0.5em{\scshape i\kern-0.25em b}\kern-0.8em\TeX}}}

\copyrightyear{2020}
\acmYear{2020}
\setcopyright{acmcopyright}\acmConference[MM '20]{Proceedings of the 28th ACM International Conference on Multimedia}{October 12--16, 2020}{Seattle, WA, USA}
\acmBooktitle{Proceedings of the 28th ACM International Conference on Multimedia (MM '20), October 12--16, 2020, Seattle, WA, USA}
\acmPrice{15.00}
\acmDOI{10.1145/3394171.3413511}
\acmISBN{978-1-4503-7988-5/20/10}

\begin{document}

\title{One-shot Text Field Labeling using Attention and Belief Propagation for Structure Information Extraction}

%
\author{Mengli Cheng, Minghui Qiu$^{*}$, Xing Shi, Jun Huang, and Wei Lin}
\affiliation{%
\authornote{The corresponding author.}
  \institution{Alibaba Group}
}
\email{{mengli.cml,minghui.qmh,xing.shix,huangjun.hj,weilin.lw}@alibaba-inc.com
}

\renewcommand{\shortauthors}{Mengli Cheng, Minghui Qiu, and et al.}
\renewcommand{\authors}{Mengli Cheng, Minghui Qiu, and et al}

\pdfcompresslevel 9

\begin{abstract}
Structured information extraction from document images usually consists of three steps: text detection, text recognition, and text field labeling. While text detection and text recognition have been heavily studied and improved a lot in literature, \textit{text field labeling} is less explored and still faces many challenges. Existing learning based methods for text labeling task usually require a large amount of labeled examples to train a specific model for each type of document. However, collecting large amounts of document images and labeling them is difficult and sometimes impossible due to privacy issues. Deploying separate models for each type of document also consumes a lot of resources. Facing these challenges, we explore one-shot learning for the text field labeling task. Existing one-shot learning methods for the task are mostly rule-based and have difficulty in labeling fields in crowded regions with few landmarks and fields consisting of multiple separate text regions. To alleviate these problems, we proposed a novel deep end-to-end trainable approach for one-shot text field labeling, which makes use of \textit{attention} mechanism to transfer the layout information between document images. We further applied \textit{conditional random field} on the transferred layout information for the refinement of field labeling. We collected and annotated a real-world one-shot field labeling dataset with a large variety of document types and conducted extensive experiments to examine the effectiveness of 
the proposed model. To stimulate research in this direction, the collected dataset and the one-shot model will be released\footnote{https://github.com/AlibabaPAI/one\_shot\_text\_labeling}.
\end{abstract}





\maketitle

\section{Introduction}
\begin{figure}[t!]
	\centering
	\includegraphics[width=1.0\columnwidth]{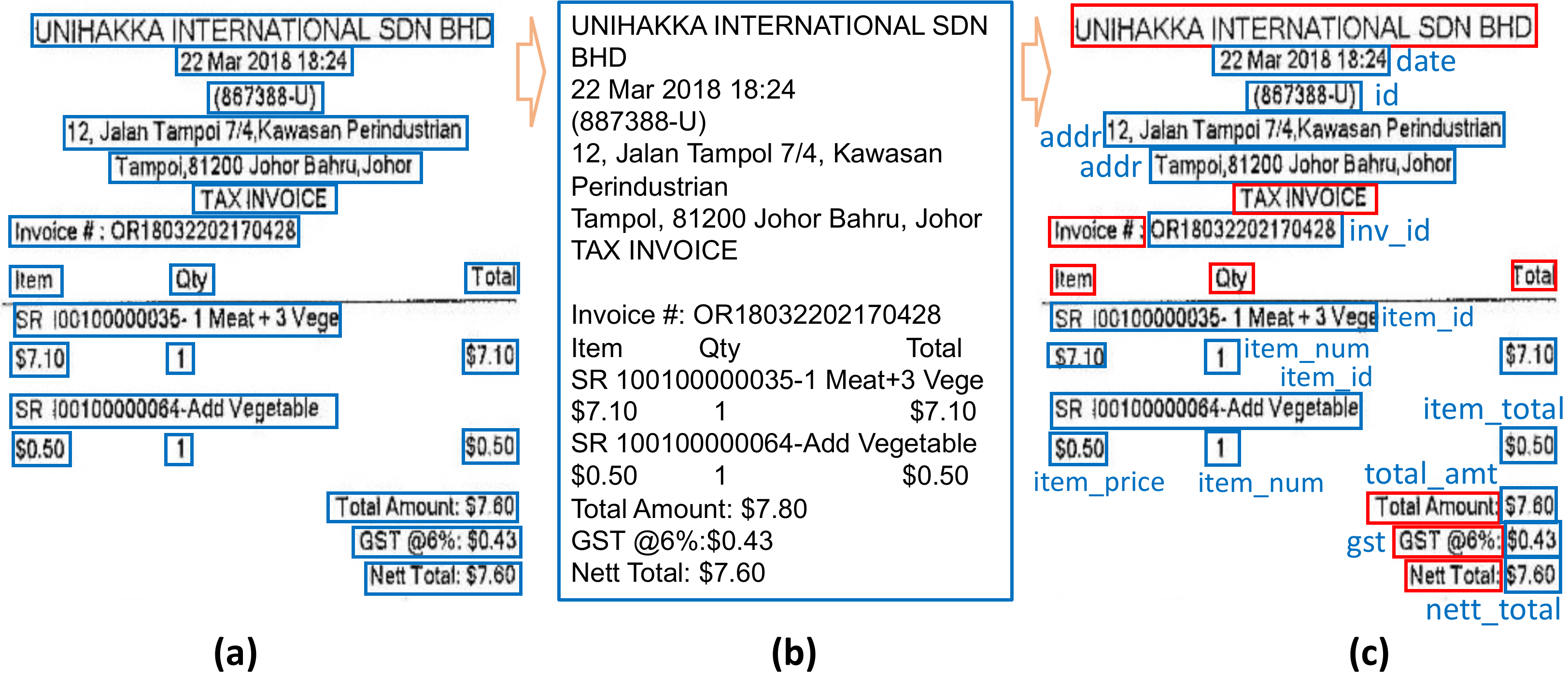}
 	\caption{Structured information extraction workflow: (a) text detection, (b) text recognition, (c) text field labeling (red rectangles are static fields, while blue are dynamic fields.) }
	\label{fig:exam}
 	\vspace{-1mm}
\end{figure}

Structured information extraction, a task to extract structured information from document images, is widely used in many industry applications, such as online user authentication and automatic filling of forms. Fig.\ref{fig:exam} shows a sample workflow of structured information extraction from a receipt image. Text detection, text recognition, and text field labeling are the three key steps for structured information extraction\cite{rusinol2013field,sunder2019one}. Text detection performance has been improved a lot due to the high performance of CNN\cite{zhou2017east}. Text recognition also achieves great progress using CTC model\cite{shi2016end} or Seq2Seq model with attention\cite{lee2016recursive}. 
With sufficient training data, it is possible to build a good text detection and recognition model for text with multiple languages in various scenes.
However, the text field labeling problem (Fig.\ref{fig:exam}c) is still less studied. The task aims to assign a label to each text region in a document so that text information could be extracted in structured formats. 

Learning based methods~\cite{chaudhury2009model,palm2018attend,yang2017learning,liu2019graph,soto2019visual} are shown to have good performance for text field labeling. 
They could automatically adapt to any type of layouts, but they usually require sufficient training data. 
As labeled examples  are  often  hard  to obtain, to alleviate this problem, we explored one-shot learning for the text field labeling task.  
Typical one-shot field labeling methods are rule based~\cite{rusinol2013field,d2018field}, which are not robust enough~\cite{d2018field} and have difficulty in extract field types consist of multiple text regions. Meanwhile, most existing one-shot methods are designed for image classification~\cite{santoro2016meta,ravi2016optimization}. Direct application of these methods on field labeling does not perform well. There are three reasons: first, text field labeling requires more fine-grained image features which is known to be challenging\cite{vinyals2016matching}; second, text fields can be with high intra-type variance (e.g. addr in Fig.\ref{fig:exam}c), third, text fields could have small inter-type variance (e.g. item\_total and total\_amt in Fig.\ref{fig:exam}c).

To alleviate the problem, we proposed an attention based deep learning method for one-shot field labeling. Our method uses a landmark to field attention approach to transfer the relationship between landmarks and fields from support document to test document. To label fields in crowded regions lack of landmarks and to label fields consisting of multiple text regions, we proposed attention based belief propagation to make use of spatial relationships between text fields to refine field labeling. 

Below we summarized our contributions.

\begin{itemize}
    \item We are the first to propose a deep end-to-end trainable approach for one-shot text field labeling problem using landmarks, attention, and belief propagation.
    \item Due to the lack of one-shot datasets, we created a dataset consisting of 43 types of documents for training and testing, and also adapted SROIE~\cite{icdar} to generate a second dataset. These two datasets will be released to stimulate research in this direction.
    \item Extensive experiments show the effectiveness of the proposed method and the usefulness of different components. Results show our method has better performance than all the competitive baseline methods. 
\end{itemize}

\section{Related Work}\label{sec:relwork}
Our work is related to text field labeling and few-shot learning, and we summarized related works in these fields in the following.
\subsection{Text field labeling.} Existing methods for the task are either learning- or rule- based.

\noindent \textbf{Learning-based methods.} Learning-based methods treat field labeling as an image labeling problem. Early methods apply markov random fields\cite{kumar2007text} or conditional random fields\cite{chaudhury2009model} to solve the problem. 
More recent approaches tend to use CNN\cite{palm2018attend,yang2017learning,soto2019visual}, GNN\cite{liu2019graph}, and Bert\cite{xu2019layoutlm} to improve performance. Despite their good performance, these methods require large amounts of training data which is hard to collect due to privacy reasons. Besides, these methods have to deploy a specific model for each type of document, which is labor-intensive and difficult to maintain. 

To reduce the requirement of large amounts of labeled data, one-shot text field labeling problem is studied. For the one-shot text field labeling problem, it is assumed the query document and support respect to the same Manhattan layout. By same layout, it means that the spatial relationships between different types of fields keep unchanged for one-type of document, such as item\_price is always on the right of the item\_name, en\_name is always under cn\_name. But there are no constraints on the horizontal distance between item\_name and item\_price or vertical distance between en\_name and cn\_name. It also does not require the document boundaries of support and query image to be well aligned. Existing one-shot or few-shot text field labeling methods are generally rule-based.

\noindent \textbf{Rule-based one-shot methods.} Medvet\cite{medvet2011probabilistic} proposed a probabilistic text sequence extraction model, which assumes simple distributions and could only extract sequences of fixed length. Recent methods tend to represent documents as attributed graphs and model it as a graph matching problem. The document graph described in \cite{peanho2012semantic} is actually a forest as each text region has at most one parent. A set of fields with fixed content is used to normalize the coordinates of query document and to establish correspondence between support and query document.  Multiple field attributes and parent-child relationships between fields have to be specified, which is difficult for end users. Match against the forest is a top-down traverse of the forest. The match process is sensitive to missing fields: mismatch of a parent field has large impacts on all its descents. 

Rusinol\cite{rusinol2013field} introduced star graph, where each field is connected to all other fields. The spatial relationship between fields are modeled by polar coordinates. A weighted voting approach is used to determine the location of center field, where the weight is calculated by tf-idf stats\cite{salton1988term}. Their approach eliminated the effort of specifying fields attributes and parent-child relationships. But the tf-idf stats are not robust enough because the satellite nodes are not pruned. 

In addition, their voting approach is sensitive to vertical displacements. The study in \cite{d2018field} alleviated the problems by referring to static fields(keywords) in the same line. But inline keywords are not available for many of the fields: item\_name and item\_price have no inline keywords. The performance degradation problem\cite{rusinol2013field} for fields in crowded regions with few static neighbors was still not solved. Hammami\cite{hammami2015one} proposed a subgraph isomorphism based method for field spotting, but their method could only work on colored forms. Existing methods generally use handcraft features and heuristics for field labeling, which are suboptimal. Moreover, all these methods have difficulty in extracting \textbf{multi-region fields} such as item\_id and item\_total(Fig.\ref{fig:exam}c), which both falls into 2 separate text regions. 

\subsection{Few-shot learning} Recent progress on few-shot learning is generally based on meta-learning\cite{santoro2016meta}. The meta-learning based methods can be grouped into three categories: metric learning, recurrent models and learning for fine-tuning. \textbf{Metric learning}. Koch\cite{koch2015siamese} proposed to use siamese networks for few-shot image recognition. Prototypical networks\cite{snell2017prototypical} learn a metric space in which classes could be represented by their centers. Relation network\cite{sung2018learning} improves prototypical networks by learning a nonlinear similarity function.    \textbf{Recurrent models}. Santoro\cite{santoro2016meta} proposed a LRU based memory augmented network for few shot learning. Finn\cite{finn2017model} improved by introducing attention mechanism into memory augmented network. While the augmented memory could generate context-aware encoding of images, the underlying LSTM network is not efficient and suffers from the order problem\cite{vinyals2015order}.   \textbf{Learning for fine-tuning}. MAML\cite{finn2017model} proposed to meta-learn good initializations of network parameters for fine-tuning on target dataset. Ravi\cite{ravi2016optimization} improved by introducing LSTM based controller to optimize the gradient descent parameters. Despite good performance, the need to fine-tune increases deployment complexity and response time.

Existing one-shot methods generally work on image classification problem. Direct application of the methods to layout labeling leads to bad performance, which is due to three reasons: firstly, labeling text contents is a fine-grained classification problem, which is challenging for existing few-shot methods\cite{vinyals2016matching}. Secondly, text contents could have high intra-class variance, such as addr(Fig.\ref{fig:exam}c). Thirdly, text contents could also be similar across class, such as item\_total and total\_amt(Fig.\ref{fig:exam}c), thus have small inter-class variance. In contrast, the spatial relationship between text fields is less variant for each type of documents. In this work, we proposed attention based approach to transfer the spatial relationship from support document to query document, which is then used for text field labeling. Our approach is end-to-end trainable and could model complex spatial relationship which is caused by multi-region fields. 

\input{model}

\input{exp}

\section{Acknowledgments}
This work was partially funded by China Postdoctoral Science Foundation (No. 2019M652038). Any opinions, findings and conclusions or recommendations expressed in this material are those of the authors and do not necessarily reflect those of the sponsor.

\bibliographystyle{ACM-Reference-Format}
\balance 
\bibliography{sdt}

\end{document}

%% file: model.tex
\section{Methodology}\label{sec:method}
Like existing one-shot labeling methods, we represented documents as graphs. Let $L$ to be the set of static text regions (called as \textbf{Landmark} later), $F$ to be the set of dynamic text regions (called as \textbf{Field} later), $LF = \{(u,v): u \in L \quad and \quad v \in F\}$, $FF = \{(u,v): u \in F \quad and \quad v \in F \}$, $Y$ be the labels of $F$, we can represent the document as a directed graph $G = \{L, F, LF, FF, Y\}$. We differentiated the two sets of text regions because the static ones could be located easily by keywords matching and be used as correspondence. Let $G_s$ and $G_q$ be the graphs defined on support document and query document, the one-shot text labeling problem is to determine labels $Y_q$ of $G_q$ given labels $Y_s$ of $G_s$. Note that the text field labeling problem could not be directly solved by graph matching\cite{li2019graph,zanfir2018deep} due to the existence of many to many matches of multi-region fields. 

\begin{figure}[th]
	\centering
	\includegraphics[width=1.0\columnwidth]{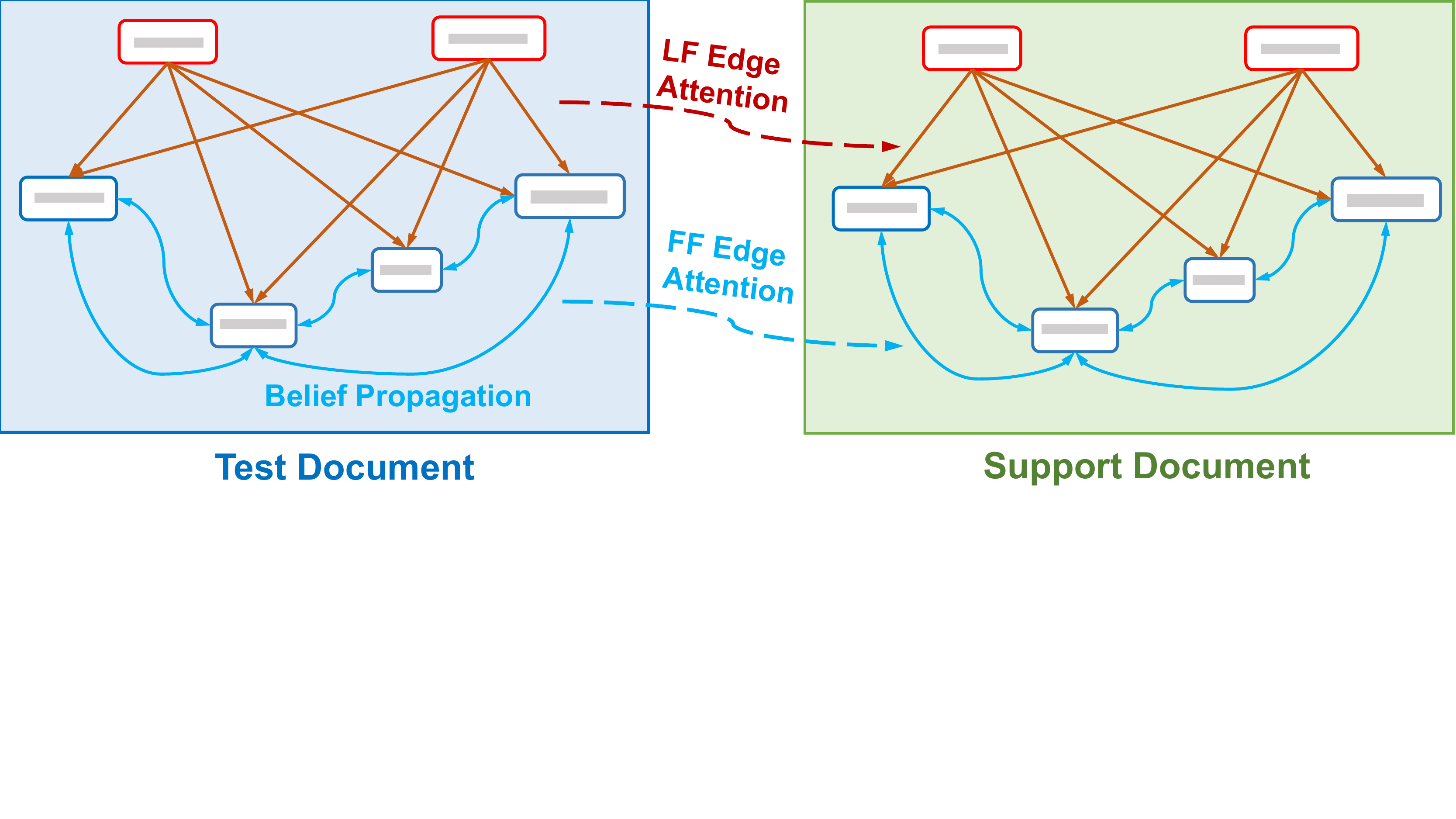}  
	\vspace{-2.5cm}
	\caption{One-shot field labeling model (Landmarks are in red, Fields in blue), including 3 components: (1) attention on LF edges, (2) attention on FF edges, (3) belief propagation.}
	\label{fig:model}
\end{figure}

Fig.\ref{fig:model} describes the overall picture of our model: first, attention is conducted on landmark to field edges(LFAttn) to transfer the spatial relationship between landmarks and fields from support document to query document and generate probabilistic distributions of fields; second, attention is conducted on field to field edges(FFAttn) to transfer the spatial relationship between fields and generate probabilistic distributions of field pairs; finally, belief propagation is applied on the two distributions to generate the final distributions of fields. 

\subsection{Document Graph Construction}
We begin our method by describing how to setup the vertices and edges for both support and query document graphs.

\noindent \textbf{Graph Vertices.}
For support document, both the static fields(L) and dynamic fields(F) on support document are specified by end users. Although there exists some automatic static regions(landmarks) detection methods~\cite{rusinol2013field,aldavert2017automatic}, they are not robust enough. In their methods, text regions of same content are taken as landmarks. Although they reduce the effort to specify landmarks, they could cause ambiguities because fields of different types could have the same content. 

For query document, the landmarks on test images can be identified as follows: first, we performed text detection\cite{yang2018inceptext} and text recognition\cite{lee2016recursive}. Then we found out landmarks by matching their text contents with those in the support documents. Finally, we could calculate rough location of characters in landmarks using attention map, and thus the position of landmarks in test image. If a text line has extra text, the quadrilateral for extra text is calculated in the same way. The detection model and recognition model used in this paper are publicly available as a service \footnote{https://ocr.data.aliyun.com/experience\#/?first\_tab=general}. 

\noindent \textbf{Graph Edges.} We put an edge between each landmark and each field. In theory, we could also put an edge between every pair of fields, but we avoided this because the setting takes too much memory. In our method, edges are only added for pairs of fields which are visible to each other. Specially, each field A emits 36 rays from its center to surrounding fields, with the angle increases by 5 degrees to previous line. A pairwise connection is added between the field A and the first fields hit by the rays of A. Finally, loop edges are added to each field so that each field could propagate a message to itself. Detailed comparison of memory consumption for full edges and sparse edges is conducted in the FFAttn section. 

\subsection{LFAttn: Attention on Landmark to Field edges}
 We applied attention mechanism to transfer the spatial relationship between landmarks and fields from support documents to query documents. The landmark to field edge features (LF features) are simply the coordinates of landmark and field pairs normalized by their joint bounding box. Similar to prototypical networks~\cite{snell2017prototypical}, the LF features in support documents are first aggregated by type to form type centers. By comparing with type centers, the aggregation step speeds up computation and improves performance\cite{snell2017prototypical}.

Let $f_L(i, j)$ be landmark to field features for $F_i$ w.r.t a landmark $L_j$. In support document, the average LF edge features of fields with label $k$ w.r.t landmark $L_j$ is defined as:
\begin{equation}
c_{kj} = \frac{1}{|\{F_i | y_i = k\}|} \sum_{y_i = k} f_L(i, j)
\end{equation}

Like relation network\cite{sung2018learning}, we used non-linear similarity score functions at attention step: the features from test field and support field are concatenated, and then MLP is applied to generate attention scores: 
\begin{equation}
\begin{aligned}
Attn(f_L(i, j), c_{kj}) &= MLP(f_L(i, j) \oplus c_{kj})
\end{aligned}
\end{equation}
The attention scores are a tensor of shape $|L| \times |F| \times K$, where $K$ is the number of field types. Note that $K$ could be a different number for different types of documents.  We then applied reduce mean operation over the landmark dimension, which generates the final scores \(S\) of shape $|F| \times K$. Finally we applied softmax cross entropy loss on the scores $S$ and the labels of fields. 

In experiments we found that it is important to average LF edge features after the attention process. The reason is that the direction of LF features is more important than the magnitude, and adding up different LF features could lose important direction information. An intuitive example is that, 
assume the vector from landmarks a/b to field i are \(LF_{ia}=(1,1), LF_{ib}=(-1,-1)\), then \(LF_{ia}+LF_{ib}=(0,0)\). The summary result (0,0) losses the original direction information. 

\subsection{FFAttn: Attention over Field to Field edges}
It is natural to consider GNN based methods\cite{liu2019graph} to make use of the relationships between fields. We conducted initial experiments with GNNs(MPAttn) in the experiment section, but the improvement is not significant. The reason is the same as that for LFAttn: aggregation of messages before attention could cause information loss. Here, we proposed to apply pairwise Conditional Random Field (CRF) to utilize the spatial relationship between fields. We presented ablation studies between GNN and our method in the experiment section.
We utilized attention mechanism to generate probabilistic distributions for FF edges. The edge features between fields (FF features) are generated in the same way as LF features and attention over the features is conducted similarly. 

Let \(FF_{k_1 k_2}\) be set of edges between fields with label \(k_1\) and fields with label \(k_2\). Let \(f_F(x_i, x_j)\) be the edge features between field i and field j. The FF edge features in support document image between fields with label \(k_1\) and fields with label \(k_2\) are defined as: 

\begin{equation}
\begin{aligned}
& c_{k_1 k_2} = \frac{1}{|FF_{k_1 k_2}|} \sum_{y_i = k_1, y_j = k_2}^{} f_F(i, j) \\
& Attn(c_{k_1 k_2}, f_F(i, j)) = MLP(c_{k_1 k_2} \oplus f_F(i, j)).
\end{aligned}
\end{equation}  

\noindent\textbf{Memory consumption reduction.} If all pairwise edges are inclued, there could be \(|F|^2\) edges between fields in each image, and the memory consumption will be $\alpha \times |F|^2 \times K^2$($\alpha$ is constant) at edge attention step, which is too heavy for modern GPUs. To avoid out of memory (OOM) issue, we considered only connections between fields that are visible to each other. As the test image and the support image are of the same layout, this approach will generate similar edge distributions, but memory consumption is reduced to $\alpha \times \beta \times |F| \times K^2$ ($\alpha$ and $\beta$ are constants). On our test set, the average edge number is 207, and the average field number is 26, so $\beta$ is approximately 8. In this case, using sparse edges helps to reduce more than 2/3 memory consumption.

\subsection{BP: belief propagation}
We considered BP for inference of pairwise CRF as it could be easily integrated into an end-to-end trainable pipeline. After FFAttn, a belief propagation matrix \(Q\) of dimension \({|F|\times |F| \times K \times K}\)  is generated. Each cell \(Q_{i, j, k, l}\) represents the joint probability of \(C_{i} = k, C_{j} = l\).  Then one standard BP step for all fields to field \(i\) could be described as
     $P_{i} = \prod_{j=0}^{N-1} Q_{ji} P_{j}$
, where $Q_{ji} P_{j}$ is the message(or belief) passed from field j to i. To enable end-to-end train, BP is done for constant number of steps, while traditional BP ends until reaching the fix point or maximal steps. The message passing along each edge is carried out in parallel in each step. As BP works together with FFAttn, we coined the term \textbf{AttnBP} to represent them thereafter.

\begin{figure}[t!]
	\centering
	\includegraphics[width=1.0\columnwidth]{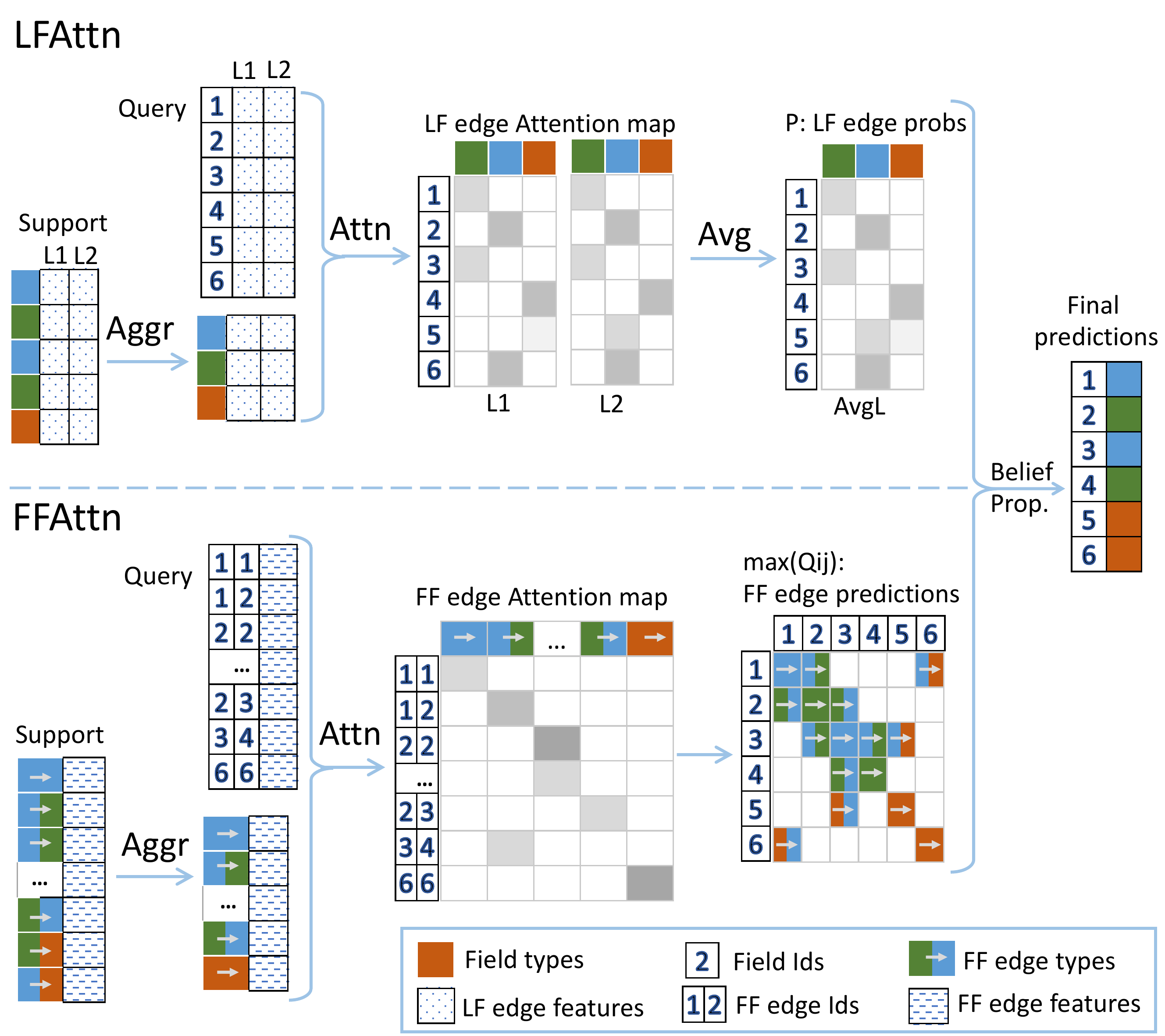}
    \caption{One-shot field labeling model details. L1/L2: landmarks. Aggr: aggregate features by field types or FF edge types. Attn: attention on edge features. Avg: average over the landmark dimension. The RGB colors represent three different field types. The degree of blackness in attention maps is proportional to similarity. Note that LFAttn generates wrong labels (predicted as green, actually red) for field 5, which is corrected by message passed from field 3.}
	\label{fig:sdt_network}
\end{figure}

We presented the details of our one-shot field labeling model in Fig~\ref{fig:sdt_network}. The up part is about LFAttn, which encodes first-order information on fields. The bottom part is about FFAttn, which encodes second-order information between fields. These two kinds of information are passed to BP module to generate the final probability distributions of fields. Note that, FFAttn could improve the results when field labels could not be fully determined by landmarks. For example,
LFAttn generates wrong labels for field 5, which is corrected by message passed from field 3. 
The up part can be trained alone or jointly with BP. The bottom part and BP can be also trained alone by replacing the output of up part with uniform label distributions. 

%% file: exp.tex
\section{Experiments}\label{sec:exp}
\subsection{Datasets}
Due to the lack of one-shot field labeling datasets, we collected and annotated a new dataset and adapted an existing dataset to form another dataset. Both  datasets will be released. 

\begin{figure}[t!]
\includegraphics[width=\linewidth]{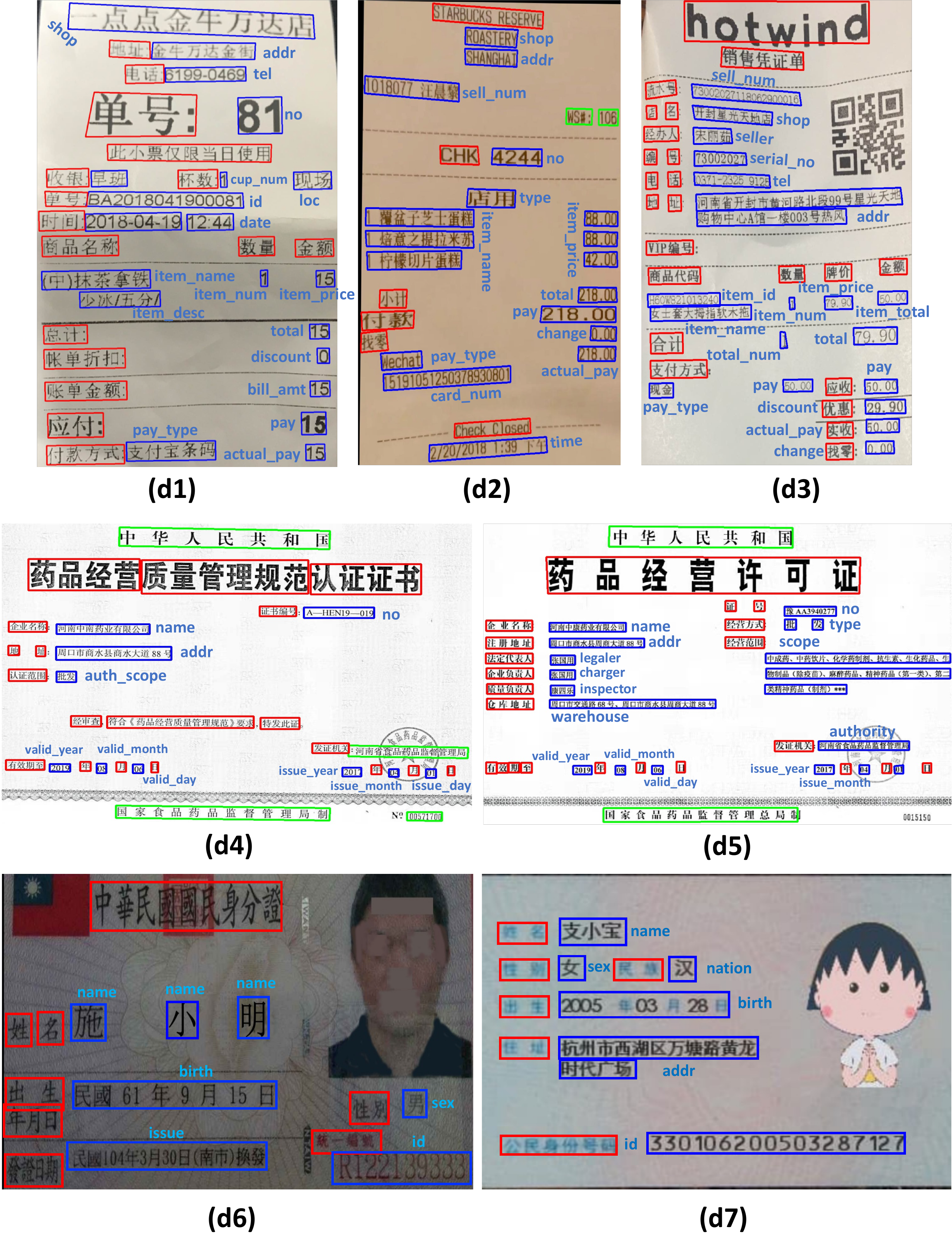}
\caption{Sample receipts (d1,d2,d3), administrative docs(d4, d5), and certificates(d6, d7). Red boxes:landmarks, blue: fields, green: background fields.}
\label{fig:sample_images}
\vspace{-0.5cm}
\end{figure}

\noindent \textbf{DocLL-oneshot.} 
We created a new dataset consisting of 1307 documents of 43 different types. 
The dataset is split into two parts, training set with 31 types and testing with 12 types.  The train set could be further grouped into 3 big categories: 15 types of receipts, 8 types of certificates and 8 types of administrative documents. The test set could also be grouped into the same big categories: 8 types of receipts, 2 types of administrative documents, and 2 types of certificates. 
Fig.\ref{fig:sample_images} shows sample images from the test set. The dataset is challenging because the document images are taken by smart phones instead of scanned, thus uneven illumination, 3D distortions\cite{ma2018docunet}, and noises are quite common. 

\begin{figure}[t!]
\includegraphics[width=\linewidth]{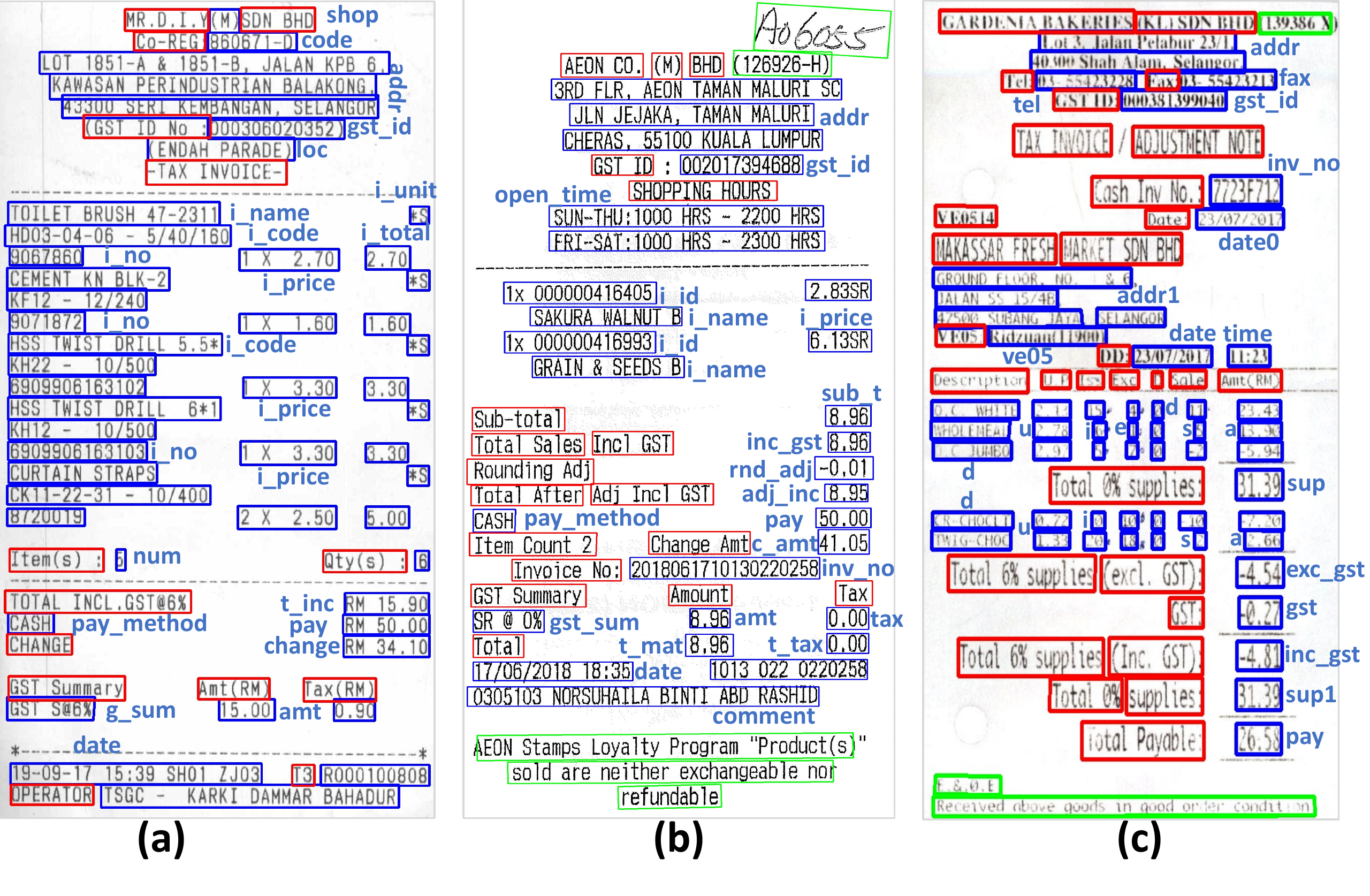}
\caption{Sample receipts from SROIE dataset. Red boxes:landmarks, blue: fields, green: background fields.}
\label{fig:sample_images_sroie}
\vspace{-0.3cm}
\end{figure}

\noindent \textbf{SROIE-oneshot.} Furthermore, we adapted the SROIE\cite{icdar} dataset to create an additional dataset. We grouped the receipts by shop names, and kept groups of more than 10 receipts, ended up with 10 kinds of receipts. 
The SROIE-oneshot will be used as an additional test set. Although the images are scanned and have less distortions, the dataset is still challenging because they also includes lots of \textbf{multi-region} fields(Fig.\ref{fig:sample_images_sroie}), and different field types are interleaved, such as i\_name, i\_price, i\_unit, i\_code, i\_no in Fig.\ref{fig:sample_images_sroie}(a). 

\noindent \textbf{Groundtruth Generation.} The two datasets used in this work are labeled by two annotators separately. After that, we crossed check the labels and fix conflicts. After labeling, each image is rotated so that the average angle of all texts is zero. Finally, the unlabeled text lines are detected by the IncepText detector\cite{yang2018inceptext} and labeled as background classes. 

\noindent\textbf{Train.} We used Tensorflow to implement our models, all the models are trained on 1080Ti with 10G memory. For all experiments, we trained with SGD using a batch size of 8 and 20000 training iterations. The momentum of SGD is 0.9, and initial learning rate is 0.01, and decayed by 0.1 after every 5000 iterations. Note that if BP is used, more GPU memory will be consumed. The training process takes around 5.5 hours which is around 2 times higher than the one without BP. For each batch, we random sampled 8 types of templates, and then random sampled a pair of images from each type of templates. The whole training process takes 2.5/5.5 hours depending on whether BP is used. If BP is used, we placed its ops and gradients on a separate gpu, which can be easily implemented using tf.device api. For inference, all modules of the model are placed on a single gpu.

\noindent\textbf{Evaluation.} For each query document Q, its support set S includes all other documents of the same class. The accuracy of Q is defined as the mean accuracy of pairs in \(\{(Q, Q') | Q' \in S\}\), which is then averaged by type. The final accuracy is defined as the mean accuracy of all types.

\subsection{Overall results}
\begin{table}[t!]
\begin{center}
\begin{tabular}{p{2.8cm} p{1.8cm} r  r} \toprule
 Method                          &  Dist                     & DocLL-      & SROIE-  \\
                                 &                           & oneshot     & oneshot \\\hline
\multicolumn{3}{l}{* Rule-based }    \\ \hline
IncSt\cite{rusinol2013field}     &   - & 0.777            & 0.764        \\ 
HybridIncSt\cite{d2018field}     &   - &  0.793           & 0.783        \\ \hline
\multicolumn{3}{l}{* Image features }    \\ \hline
Siamese     & Euclidean          & 0.414                  & 0.257     \\
MatchNet    & Euclidean          & 0.665                  & 0.593     \\
ProtoNet    & Euclidean          & 0.638                  & 0.550     \\
RelNet      & Learned            & 0.627                  & 0.566     \\
\hline 
\multicolumn{3}{l}{* LF Features}      \\  \hline 
Siamese     & Cosine             & 0.328                  & 0.206            \\
MatchNet    & Cosine$\dagger$    & 0.941                  & 0.969            \\
ProtoNet    & Cosine             & 0.938                  & 0.960            \\
Ours        & Learned            & 0.939                  & 0.965            \\
\hline
   \multicolumn{3}{l}{* LF Features + FF Features}  \\ \hline
MatchNet    & Cosine$\dagger$   & 0.942                & 0.970             \\
ProtoNet    & Cosine            & 0.948                & 0.973             \\
Ours        & Learned           & \textbf{0.964}       & \textbf{0.985}    \\
\bottomrule
\end{tabular}  
\caption{Comparison with state of art one-shot methods. $\dagger$ means we added linear transform to Cosine distance.}
\label{tabl:cmp_exist}
\end{center}
\vspace{-0.5cm}
\end{table}

In this section, we compared our method with existing rule-based, image feature based, and landmark feature based methods.
The existing one-shot methods using text image features are based on VGG like networks\cite{koch2015siamese}, but modern CNN networks with residual connections are more accurate and efficient. Hence in this work, a Resnet-18 variant that has better performance is used for feature extraction.  
We also extended existing rule-based methods(IncSt\cite{rusinol2013field}, HybridIncSt\cite{d2018field}) in order to handle \textbf{multi-region} types of fields: we did not vote for the most probable text region for each type, we voted for the most probable type for each text region instead. 

The results are presented in Table~\ref{tabl:cmp_exist}. First, existing few-shot methods on image features have lower performance than rule-based methods. The reason is that image features of different types are not easy to separate as intra-type variance is large compared with inter-type variance for many field types. In contrast, the spatial relationship between fields are relatively stable, thus their intra-type variance is small compared with inter-type variance. Our LFAttn method improves the rule-based methods, as the learned features maximize the gap between different types and generate more robust classifiers than handcraft rules. Besides, in contrast to rule-based methods that extract different field types separately, our methods extract all fields types at once. We could then take into consideration of spatial constraints between fields by making use of AttnBP, which further improves the results by $\sim3\%$ (from 0.939 to 0.964). Second, for MatchNet and ProtoNet with landmark features, we used cosine similarity instead of euclidean similarity. The reason is that LF features do not conform to exponential distribution assumed by euclidean distance\cite{snell2017prototypical}. The variance in magnitude could be large due to the difference in image sizes, while the variance in angles is relatively small. Last, the FF feature distribution is more complex than LF feature distribution due to the existence of \textbf{multi-region} field types. When FF features is used, ProtoNet has better performance than MatchNet as it poses attention on the type center which is more stable~\cite{snell2017prototypical}. Our method uses learned nonlinear similarity function, which is shown to improve the ProtoNet. 

In a nutshell, with the incorporation of attention (LFAttn) and belief propagation (AttnBP), our method achieves the state-of-the-art results on all the datasets.

\begin{table}[t!]
\begin{center}
\begin{tabular}{l c  c  c  c} \toprule
               Methods   &       All\_L             &   AttnBP             & \multicolumn{2}{c}{AttnBP}   \\\cline{4-5}
                         &                          &                      &   +All\_L    &   +Det\_L     \\
\midrule 
DocLL-oneshot           &  0.936   &   0.934   &  \textbf{0.964}      & 0.961  \\
SROIE-oneshot            &  0.960   &   0.962   &  \textbf{0.985}      & 0.984  \\ \bottomrule
\end{tabular}  
\caption{The importance of landmarks and BP. All\_L: LFAttn use all landmarks, Det\_L: LFAttn use landmarks detected in prepossessing step.}
\label{tab:sdtacc}
\end{center}
\end{table}

\subsection{The importance of landmarks and BP}
Here we conducted model ablation studies to examine the importance of landmarks and belief propagation (BP).
As shown by Table \ref{tab:sdtacc}, the results on both test datasets are consistent. A comparison of using All\_L+AttnBP and AttnBP shows that landmarks contribute to an increase of 3\% of accuracy on average. For belief propagation, two message passing steps are used as more steps leads to no more improvement. The introducing of attention based belief propagation module greatly improves the labeling accuracy. It also shows that AttnBP itself can achieve very good performance. 

\begin{figure}[t!]
	\vspace{-3mm}
	\centering
	\includegraphics[width=1\columnwidth]{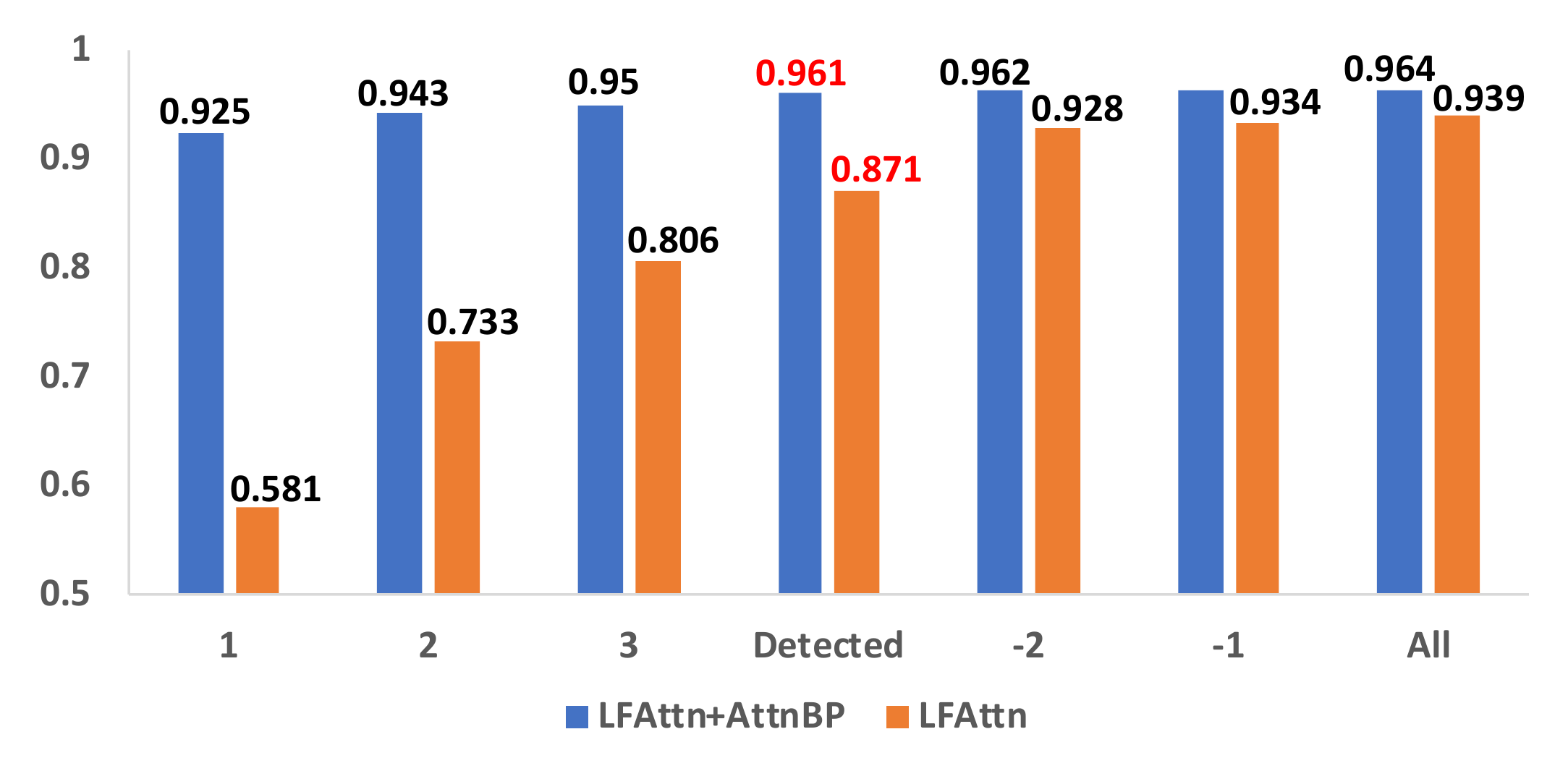}
	\caption{Document field labeling accuracy vs number of landmarks. All:use all landmarks, -1/-2:random drop 1/2 landmarks, detected:using landmarks from prepossessing step, 1/2/3: only keep 1/2/3 landmarks.}
	\label{fig:anchor_num}
	\vspace{-0.2cm}
\end{figure}

We further examine the impacts of the number of landmarks on labeling accuracy in Fig.\ref{fig:anchor_num}. As expected, the labeling accuracy grows as the number of landmarks increases. 
The overall accuracy is good if dropping less than 3 landmarks (see -1,-2 in x-axis). 
Our existing landmark detection method misses 1.7 landmarks in average, but with a large variance (the maximal number of landmarks missed is 16), 
so the final performance is worse than randomly dropping 2 landmarks. The reason is that our text detection and recognition models are not specifically trained for documents, hence maybe not very precise in landmark detection.
If we collected more documents to finetune those models, this issue can be alleviated. Another way to alleviate this is to use end-to-end training with AttnBP. As in Fig.\ref{fig:anchor_num}, the accuracy of LFAttn+AttnBP is less affected by the number of landmarks because AttnBP introduces second-order information. Thus, even some of the landmarks are not detected, the accuracy is almost the same. Moreover, AttnBP could handle complex table layout which is composed of interleaved field types in crowded regions where fewer landmarks are available. More details are presented in the case-study section. 


\begin{table}[t!]
\begin{center}
\begin{tabular}{l  c  c} \toprule
Methods          & DocLL-oneshot  & SROIE-oneshot   \\ \hline
AvgAttn          &    0.774        & 0.702           \\
LFAttn           & \textbf{0.939}  & \textbf{0.965}  \\
\hline
LFAttn + MPAttn  & 0.946           & 0.978          \\ 
LFAttn + GAT     & 0.948           & 0.979          \\
LFAttn + AttnBP  & \textbf{0.964}  & \textbf{0.985} \\ \bottomrule
\end{tabular}  
\caption{Comparison of various attention methods. }
\label{tabl:cmp_attn}
\end{center}
\end{table}

\subsection{The effectiveness of attention}\label{sec:attn_exp}
Our model achieves high accuracy on a variety of test documents which are not seen in training phase. It shows that layout information transferred by attention mechanism is effective. We further tested two alternative designs to verify the effectiveness of the design of attention module:

\noindent \textbf{Average before Attention (AvgAttn).}
The average operation over the landmark dimension is moved to the front of attention operation. As in Table \ref{tabl:cmp_attn}, the accuracy is far behind that of LFAttn. We also found reduce-max has no improvement. Without attention, aggregation over LF features mixes up the information from all landmarks and leads to less distinguishable features. 

\noindent \textbf{Message Passing before Attention (MPAttn).} Following the graph matching network~\cite{li2019graph}, we considered message passing before field feature attention, which is defined as: 
\begin{equation}
\begin{aligned}
      f(i) &= \sum_{j=1}^{|F|} MLP(f_F(i, j) \oplus Pad(P_i) \oplus Pad(P_j)) \\
      c_k &= \frac{1}{|\{F_i | y_i = k\}|} \sum_{y_i = k} f(i)
\end{aligned}
\end{equation}
where $P_i$ is from LFAttn, and \(Pad\) is to pad to the max number of types (48 in experiments). The field feature $f(i)$  of support document is built in the same way except that $P_i$ is the one hot encoding of $y_i$. After message passing, attention of $f(i)$ over $c_k$ is conducted in the same way as LFAttn. Finally, residual structure is applied to improve performance; otherwise, the performance is worse than LFAttn. As shown in Table~\ref{tabl:cmp_attn}, the improvement is less than AttnBP. We also tried graph attention networks(GAT)\cite{velivckovic2017graph}, the improvement is also inferior to our methods. With attention ahead of message passing, the message types can be determined first, and messages passed between fields are more informative.

\subsection{Case study}
In this section, we presented a case study on document d0 - one of the most challenging documents in our dataset. 
The accuracies for d0 are: HybridIncSt 0.72, RelNetImg 0.77, LFAttn 0.84, and LFAttn+AttnBP 0.93. Our LFAttn+AttnBP method performs consistently better than all the other methods. 

\begin{figure}[t!]
    \includegraphics[width=\linewidth]{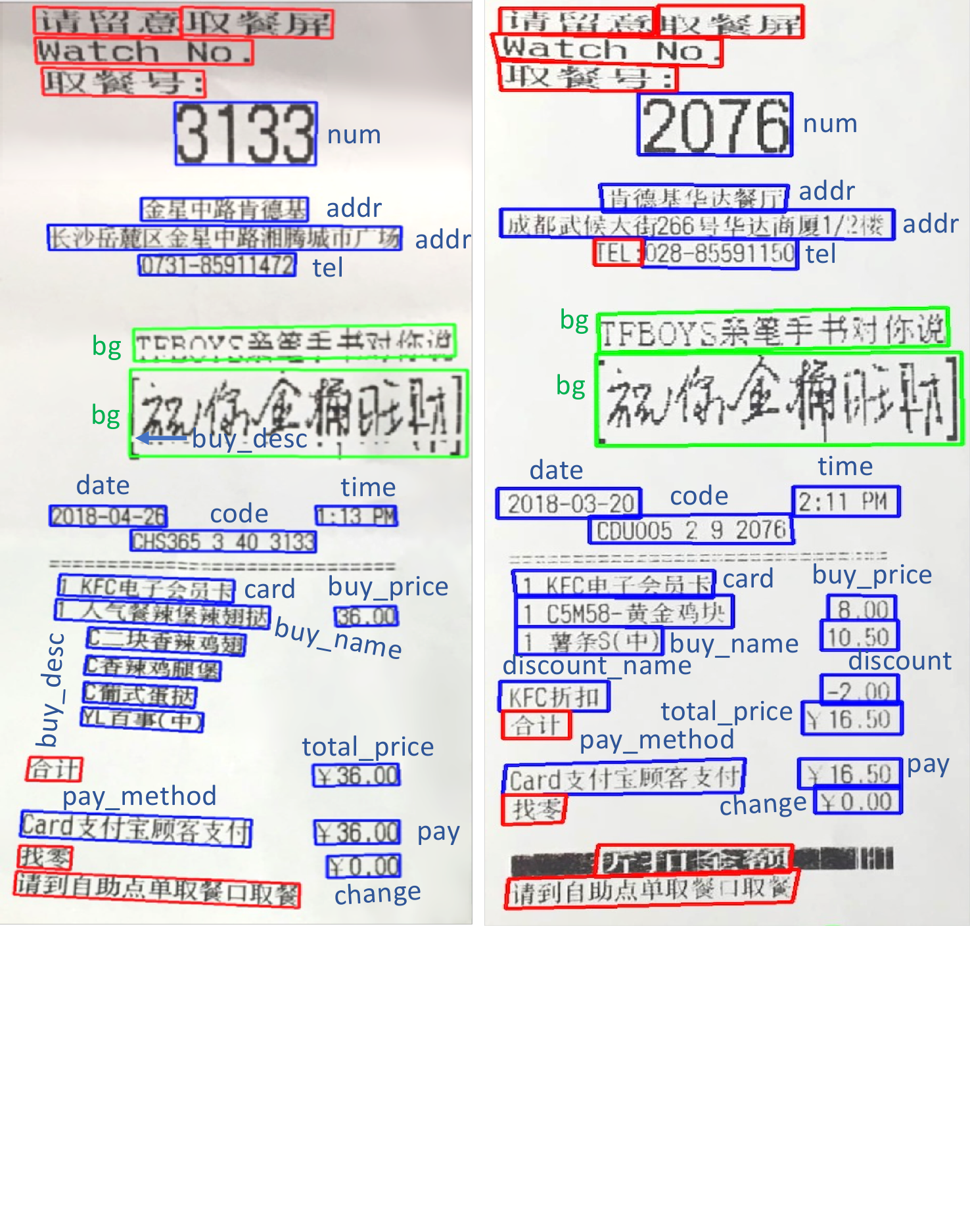}
    \vspace{-3.0cm}
    \caption{Sample d0 from one-shot dataset.}
    \label{fig:sample_images_d0}
\end{figure}

\definecolor{word_red}{RGB}{255,0,0}
\definecolor{word_purple}{RGB}{112,48,160}
\definecolor{word_green}{RGB}{0,176,80}
\definecolor{word_blue}{RGB}{0,176,240}
\definecolor{word_brown}{RGB}{197,90,17}

\begin{figure*}[t!]
	\centering
	\includegraphics[width=2.1\columnwidth]{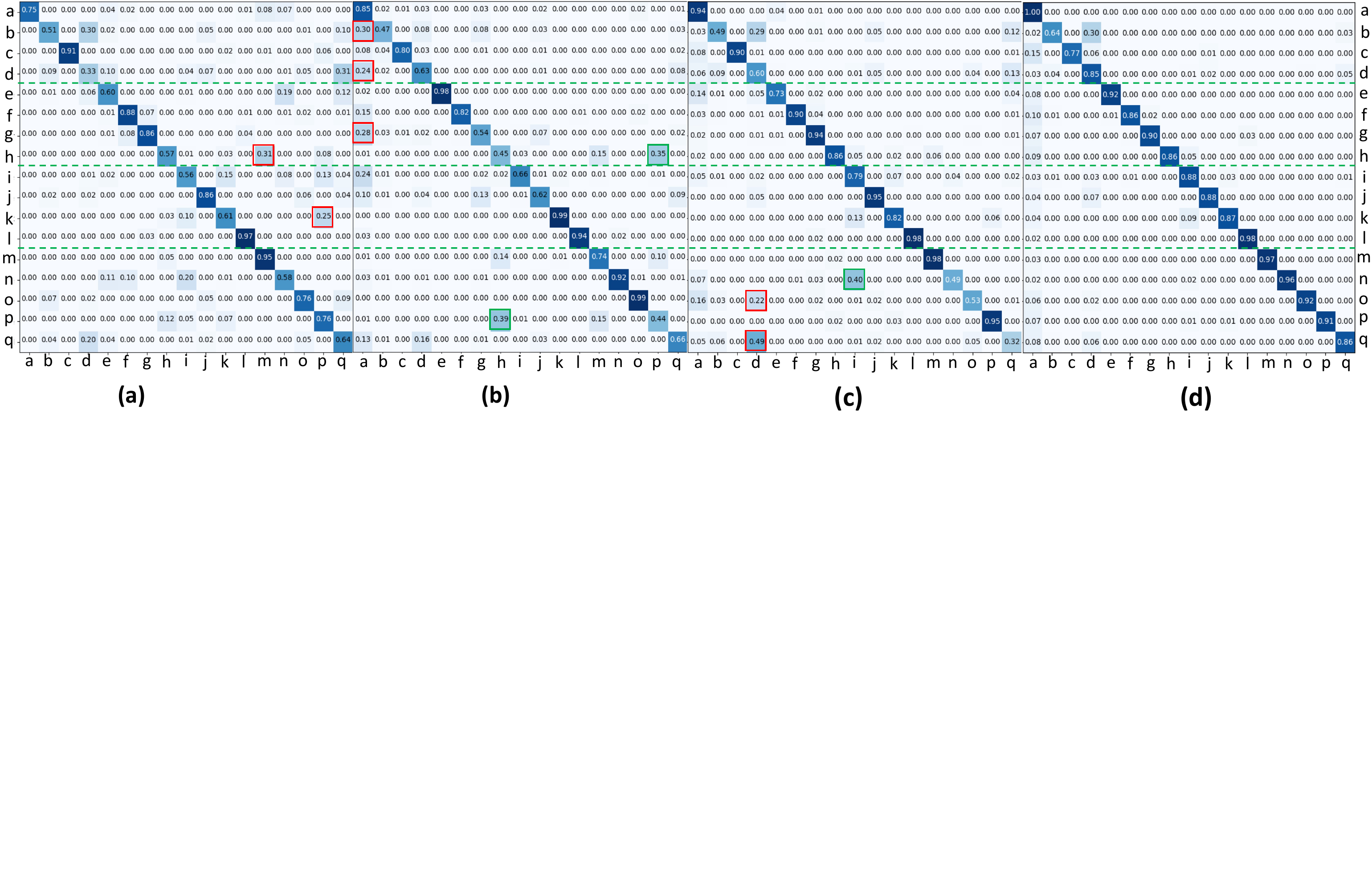}
	\vspace{-6.5cm}
	\caption{Confusion matrix for document d0~(best viewed in color). x axis: prediction, y axis: groundtruth.(a) HybridIncSt, (b) RelNetImg, (c) LFAttn, (d) LFAttn+AttnBP. Field label alias: a:background, b:buy\_desc, c:pay\_type, d:buy\_name, e:code, f:tel, g:addr, h:pay, i:buy\_price, j:discount\_name, k:discount, l:num, m:change, n:time, o:date, p:total\_price, q:card.}
	\label{fig:all_confusion}
\end{figure*}

To dive into the details of our model performance, we presented the confusion matrix of HybridIncSt, RelNetImg, LFAttn, and LFAttn+ AttnBP for document d0  in Fig.\ref{fig:all_confusion}. Rule-based method (Fig.\ref{fig:all_confusion}a) has difficulty in classifying m:change and h:pay(\textcolor{word_red}{$\square$}), as well as k:discount and p:total\_price(\textcolor{word_red}{$\square$}), as they are distant from landmarks. The document is challenging for RelNetImg (Fig.\ref{fig:all_confusion}b) in two aspects: first, the text contents of g:addr, b:buy\_desc, d:buy\_name fields are highly variant, thus many of them are misclassified as a:background(\textcolor{word_red}{$\square$}); second, fields of h:pay, p:total\_price, and m:change(\textcolor{word_green}{$\square$}) have similar content.
The document is also challenging for LFAttn(Fig.\ref{fig:all_confusion}b) because there are lots of multi-region fields in the middle but there are few landmarks in the middle: the confusion between d:buy\_name, o:date, and q:card is high(\textcolor{word_red}{$\square$}), so as the confusion between i:buy\_price and n:time(\textcolor{word_green}{$\square$}). A comparison of the confusion matrix(Fig.\ref{fig:all_confusion}c, Fig.\ref{fig:all_confusion}d) shows that LFAttn+AttnBP clears the above confusions. 

\noindent \textbf{Per Document Type Results.} As shown in Table~\ref{tabl:per_type_acc}, our approach has the best performance for almost all types of document. It can also be seen that receipts are usually more challenging than administrative documents and certificates, because receipt layouts are more complex and have fewer landmarks. Rule-based methods perform good for fields that are close to landmarks(d3, d6, d7); but they have low performance on documents of complex(d1, d2). They also perform bad on d4, d5 because they are \textbf{sensitive to background(bg) fields}: many background fields are labeled incorrectly as issue\_* and valid\_* because the position of issue\_* and valid\_* w.r.t to landmarks could shift. If the background fields are dropped, their performance improves a lot~(d4:0.997, d5:0.954). As the percentages of background fields are relatively large(17\% for DocLL-oneshot test dataset and 16\% for SROIE-oneshot dataset), we further evaluated the impact of background fields on labeling accuracy by dropping all the background fields: Table~\ref{tabl:sensitive_noise} indicates that rule based methods are sensitive to background fields, while our method is almost not affected by this issue, thus our method is more robust.

\begin{table}[t!]
\begin{center}
\begin{tabular}{l r r r  r r r r} \toprule 
Methods     &   d1     &   d2   &  d3    &  d4   & d5     & d6    & d7     \\ \hline
   \multicolumn{4}{l}{* Rule based }       \\ \hline
IncSt  &  0.83    & 0.72   & 0.91   & 0.69  &  0.75  & 0.96  & 0.91   \\           
HybridIncSt   &  0.84    & 0.73   & 0.92   & 0.71  &  0.80  & 0.96  & 0.99   \\ \hline    
  \multicolumn{4}{l}{* Image features }        \\ \hline
MatchNet &  0.52    & 0.65   & 0.66   & 0.76  &  0.68  & 0.65  & 0.84   \\           
ProtoNet &  0.51    & 0.64   & 0.58   & 0.71  &  0.65  & 0.65  & 0.77   \\           
RelNet   &  0.50    & 0.64   & 0.58   & 0.67  &  0.62  & 0.67  & 0.83   \\ \hline    
   \multicolumn{4}{l}{* LF Features}      \\  \hline 
MatchNet &  0.91    & 0.90   & 0.98   & 1.00  & 0.97   & 0.97  & 1.00   \\           
ProtoNet &  0.90    & 0.88   & 0.98   & 1.00  & 0.99   & 0.98  & 0.99   \\ 
Ours   &  0.91    & 0.88   & 0.98   & 1.00  & 1.00   & 0.98  & 1.00   \\ \hline  
   \multicolumn{4}{l}{* LF Features + FF features}  \\ \hline
MatchNet &  0.91    & 0.90   & 0.99   & 1.00  & 0.98   & 0.97  & 1.00   \\           
ProtoNet &  0.90    & 0.94   & 0.99   & 0.98  & 1.00   & 0.96  & 0.99   \\ 
Ours   &  0.93    & 0.95   & 0.99   & 1.00  & 1.00   & 0.97  & 1.00   \\ \bottomrule
\end{tabular}
\caption{Per-type labeling accuracy of DocLL-oneshot test documents.}
\label{tabl:per_type_acc}
\end{center}
\vspace{-0.1cm}
\end{table}

\begin{table}[t!]
\begin{tabular}{l    c     c    c    c  } \toprule 
\multirow{2}{*}{Dataset}&  \multirow{2}{*}{IncSt}  &  \multirow{2}{*}{HybridIncSt} &  \multirow{2}{*}{LFAttn} & LFAttn    \\   
               &          &             &         &  + AttnBP \\ \midrule
DocLL-oneshot  &  0.883   &   0.891     &  0.944  &    0.971        \\
Incre          &  0.106   &    0.098    &  0.005  &    0.007        \\ \hline
SROIE-oneshot  &  0.862   &   0.866     &  0.964  &    0.985        \\
Incre          &  0.098   &   0.083     &  -0.001 &    0.000        \\ \bottomrule
\end{tabular}
\caption{The impact of background fields(bg fields) on labeling accuracy. Incre is the increase in accuracy if background fields is dropped. }
\label{tabl:sensitive_noise}
\vspace{-0.1cm}
\end{table}

\begin{table}[th!]
\begin{center}
\begin{tabular}{l p{1.4cm} r  r} \toprule
Methods  &     Dist. &     DocLL-5-shot       & SROIE-5-shot  \\ \hline
   \multicolumn{4}{l}{* Rule based }       \\ \hline
IncSt       &      -       &         0.776(-0.001)        &     0.779(+0.015) \\
HybridIncSt &      -       &         0.806(+0.013)       &     0.797(+0.014) \\  \hline
 \multicolumn{3}{l}{* Image Features} \\ \hline
Siamese  &      Euclidean &         0.435 (+0.031)       & 0.266 (+0.009)    \\
MatchNet &      Euclidean &         0.724 (+0.059)       & 0.651 (+0.058)    \\
ProtoNet &      Euclidean &         0.732 (+0.094)       & 0.651 (+0.101)    \\
RelNet   &      Learned   &         0.711 (+0.088)       & 0.663 (+0.097)    \\ \hline 
 \multicolumn{3}{l}{* Landmark Features} \\ \hline
Siamese  &      Cosine  &         0.314 (-0.014)       & 0.229 (-0.023)   \\ 
MatchNet &      Cosine$\dagger$  &         0.962 (+0.021)       & 0.983 (+0.014)   \\
ProtoNet &      Cosine  &         0.963 (+0.025)       & 0.975 (+0.015)   \\
LFAttn   &      Learned &         0.961 (+0.022)       & 0.979 (+0.014)   \\
\hline
 \multicolumn{3}{l}{* Landmark Features + AttnBP}  \\ \hline
MatchNet &      Cosine$\dagger$ &       0.952 (+0.038)               & 0.978 (+0.024)      \\
ProtoNet &      Cosine &       0.971 (+0.023)               & 0.986 (+0.013)      \\
LFAttn   &      Learned &      \textbf{0.978} (+0.014)         & \textbf{0.991} (+0.006)   \\
\bottomrule
\end{tabular}  
\caption{Comparison with the state of art five-shot methods. The numbers in () are improvements from one-shot methods. $\dagger$ means we added linear transform to Cosine distance.}
\label{tabl:cmp_exist_five_shot}
\end{center}
\vspace{-0.5cm}
\end{table}

\definecolor{word_red}{RGB}{255,0,0}
\definecolor{word_purple}{RGB}{112,48,160}
\definecolor{word_green}{RGB}{0,176,80}
\definecolor{word_blue}{RGB}{0,176,240}
\definecolor{word_brown}{RGB}{197,90,17}

\section{Few-shot Field Labeling}
Previous studies on image classification problem report that few-shot learning methods improve over their corresponding one-shot versions. We extended our one-shot approach to few-shot to further improve the performance. 

Due to the difference in scales of different images, taking the average of LF and FF features over the support set \(S\) does not lead to good results. Hence following matching networks, we took the average of the one-shot probabilities of \(|S|\) query-support pairs as the output of few-shot methods. Following the existing studies, we considered 5-shot layout labeling, i.e. $|S|=5$. 

As expected, all five-shot methods except Siamese Network perform better than one-shot methods (Table \ref{tabl:cmp_exist_five_shot}). Our LFAttn+AttnBP model consistently outperforms all other methods and outperforms image features based methods by a large margin. As the improvement of five-shot LFAttn+AttnBP over the one-shot version is relatively small (1.4\%), we took a close examination of per document type improvement and found that five-shot improves more over the challenging types: 2.6\% over 6 types of hardest documents, 0.3\% for the rest 6 types whose accuracy is close to 1(99.2\%). 

\section{Conclusions}\label{sec:conclusion}
In this work, we proposed a one-shot approach for text field labeling using attention mechanism and belief propagation. 
To the best of our knowledge, we are the first to propose a deep end-to-end trainable approach for one-shot text field labeling. Our method could handle complex document layout with multi-region fields and shows to have good performance on the evaluation datasets.

Note that our method relies on the topological relations between text regions, which is suitable for data with similar layouts. Thus the method may not have good performance for data with different layouts. 
We seek to handle such variations to improve the robustness of our method in future. Meanwhile, our method can be extended to zero-shot scenarios as well, we leave it as another future work.

%% file: sdt.bbl

\begin{thebibliography}{32}


\ifx \showCODEN    \undefined \def \showCODEN     #1{\unskip}     \fi
\ifx \showDOI      \undefined \def \showDOI       #1{#1}\fi
\ifx \showISBNx    \undefined \def \showISBNx     #1{\unskip}     \fi
\ifx \showISBNxiii \undefined \def \showISBNxiii  #1{\unskip}     \fi
\ifx \showISSN     \undefined \def \showISSN      #1{\unskip}     \fi
\ifx \showLCCN     \undefined \def \showLCCN      #1{\unskip}     \fi
\ifx \shownote     \undefined \def \shownote      #1{#1}          \fi
\ifx \showarticletitle \undefined \def \showarticletitle #1{#1}   \fi
\ifx \showURL      \undefined \def \showURL       {\relax}        \fi
\providecommand\bibfield[2]{#2}
\providecommand\bibinfo[2]{#2}
\providecommand\natexlab[1]{#1}
\providecommand\showeprint[2][]{arXiv:#2}

\bibitem[\protect\citeauthoryear{Aldavert, Rusi{\~n}ol, and Toledo}{Aldavert
  et~al\mbox{.}}{2017}]%
        {aldavert2017automatic}
\bibfield{author}{\bibinfo{person}{David Aldavert},
  \bibinfo{person}{Mar{\c{c}}al Rusi{\~n}ol}, {and} \bibinfo{person}{Ricardo
  Toledo}.} \bibinfo{year}{2017}\natexlab{}.
\newblock \showarticletitle{Automatic static/variable content separation in
  administrative document images}. In \bibinfo{booktitle}{\emph{International
  Conference on Document Analysis and Recognition}}, Vol.~\bibinfo{volume}{1}.
\newblock


\bibitem[\protect\citeauthoryear{Chaudhury, Jindal, and Roy}{Chaudhury
  et~al\mbox{.}}{2009}]%
        {chaudhury2009model}
\bibfield{author}{\bibinfo{person}{Santanu Chaudhury}, \bibinfo{person}{Megha
  Jindal}, {and} \bibinfo{person}{Sumantra~Dutta Roy}.}
  \bibinfo{year}{2009}\natexlab{}.
\newblock \showarticletitle{Model-guided segmentation and layout labelling of
  document images using a hierarchical conditional random field}. In
  \bibinfo{booktitle}{\emph{ICPRAI}}. Springer.
\newblock


\bibitem[\protect\citeauthoryear{d'Andecy, Hartmann, and Rusinol}{d'Andecy
  et~al\mbox{.}}{2018}]%
        {d2018field}
\bibfield{author}{\bibinfo{person}{Vincent~Poulain d'Andecy},
  \bibinfo{person}{Emmanuel Hartmann}, {and} \bibinfo{person}{Mar{\c{c}}al
  Rusinol}.} \bibinfo{year}{2018}\natexlab{}.
\newblock \showarticletitle{Field extraction by hybrid incremental and a-priori
  structural templates}. In \bibinfo{booktitle}{\emph{International Workshop on
  Document Analysis Systems}}.
\newblock


\bibitem[\protect\citeauthoryear{Finn, Abbeel, and Levine}{Finn
  et~al\mbox{.}}{2017}]%
        {finn2017model}
\bibfield{author}{\bibinfo{person}{Chelsea Finn}, \bibinfo{person}{Pieter
  Abbeel}, {and} \bibinfo{person}{Sergey Levine}.}
  \bibinfo{year}{2017}\natexlab{}.
\newblock \showarticletitle{Model-agnostic meta-learning for fast adaptation of
  deep networks}. In \bibinfo{booktitle}{\emph{International Conference on
  Machine Learning}}.
\newblock


\bibitem[\protect\citeauthoryear{Hammami, H{\'e}roux, et~al\mbox{.}}{Hammami
  et~al\mbox{.}}{2015}]%
        {hammami2015one}
\bibfield{author}{\bibinfo{person}{Maroua Hammami}, \bibinfo{person}{Pierre
  H{\'e}roux}, {et~al\mbox{.}}} \bibinfo{year}{2015}\natexlab{}.
\newblock \showarticletitle{One-shot field spotting on colored forms using
  subgraph isomorphism}. In \bibinfo{booktitle}{\emph{International Conference
  on Document Analysis and Recognition}}.
\newblock


\bibitem[\protect\citeauthoryear{Koch, Zemel, and Salakhutdinov}{Koch
  et~al\mbox{.}}{2015}]%
        {koch2015siamese}
\bibfield{author}{\bibinfo{person}{Gregory Koch}, \bibinfo{person}{Richard
  Zemel}, {and} \bibinfo{person}{Ruslan Salakhutdinov}.}
  \bibinfo{year}{2015}\natexlab{}.
\newblock \showarticletitle{Siamese neural networks for one-shot image
  recognition}. In \bibinfo{booktitle}{\emph{International Conference on
  Machine Learning workshop}}, Vol.~\bibinfo{volume}{2}.
\newblock


\bibitem[\protect\citeauthoryear{Kumar, Gupta, et~al\mbox{.}}{Kumar
  et~al\mbox{.}}{2007}]%
        {kumar2007text}
\bibfield{author}{\bibinfo{person}{Sunil Kumar}, \bibinfo{person}{Rajat Gupta},
  {et~al\mbox{.}}} \bibinfo{year}{2007}\natexlab{}.
\newblock \showarticletitle{Text extraction and document image segmentation
  using matched wavelets and MRF model}.
\newblock \bibinfo{journal}{\emph{IEEE Transactions on Image Processing}}
  (\bibinfo{year}{2007}).
\newblock


\bibitem[\protect\citeauthoryear{Lee and Osindero}{Lee and Osindero}{2016}]%
        {lee2016recursive}
\bibfield{author}{\bibinfo{person}{Chen-Yu Lee} {and} \bibinfo{person}{Simon
  Osindero}.} \bibinfo{year}{2016}\natexlab{}.
\newblock \showarticletitle{Recursive recurrent nets with attention modeling
  for ocr in the wild}. In \bibinfo{booktitle}{\emph{IEEE Conference on
  Computer Vision and Pattern Recognition}}.
\newblock


\bibitem[\protect\citeauthoryear{Li, Gu, et~al\mbox{.}}{Li
  et~al\mbox{.}}{2019}]%
        {li2019graph}
\bibfield{author}{\bibinfo{person}{Yujia Li}, \bibinfo{person}{Chenjie Gu},
  {et~al\mbox{.}}} \bibinfo{year}{2019}\natexlab{}.
\newblock \showarticletitle{Graph Matching Networks for Learning the Similarity
  of Graph Structured Objects}.
\newblock \bibinfo{journal}{\emph{International Conference on Machine
  Learning}} (\bibinfo{year}{2019}).
\newblock


\bibitem[\protect\citeauthoryear{Liu, Gao, et~al\mbox{.}}{Liu
  et~al\mbox{.}}{2019}]%
        {liu2019graph}
\bibfield{author}{\bibinfo{person}{Xiaojing Liu}, \bibinfo{person}{Feiyu Gao},
  {et~al\mbox{.}}} \bibinfo{year}{2019}\natexlab{}.
\newblock \showarticletitle{Graph Convolution for Multimodal Information
  Extraction from Visually Rich Documents}.
\newblock \bibinfo{journal}{\emph{The North American Chapter of the Association
  for Computational Linguistics}} (\bibinfo{year}{2019}).
\newblock


\bibitem[\protect\citeauthoryear{Ma, Shu, et~al\mbox{.}}{Ma
  et~al\mbox{.}}{2018}]%
        {ma2018docunet}
\bibfield{author}{\bibinfo{person}{Ke Ma}, \bibinfo{person}{Zhixin Shu},
  {et~al\mbox{.}}} \bibinfo{year}{2018}\natexlab{}.
\newblock \showarticletitle{Docunet: document image unwarping via a stacked
  U-Net}. In \bibinfo{booktitle}{\emph{IEEE Conference on Computer Vision and
  Pattern Recognition}}.
\newblock


\bibitem[\protect\citeauthoryear{Medvet, Bartoli, and Davanzo}{Medvet
  et~al\mbox{.}}{[n.d.]}]%
        {medvet2011probabilistic}
\bibfield{author}{\bibinfo{person}{Eric Medvet}, \bibinfo{person}{Alberto
  Bartoli}, {and} \bibinfo{person}{Giorgio Davanzo}.}
  \bibinfo{year}{[n.d.]}\natexlab{}.
\newblock \showarticletitle{A probabilistic approach to printed document
  understanding}.
\newblock \bibinfo{journal}{\emph{International Journal on Document Analysis
  and Recognition}} \bibinfo{volume}{14}, \bibinfo{number}{4}
  (\bibinfo{year}{[n.\,d.]}).
\newblock


\bibitem[\protect\citeauthoryear{on~Document~Analysis and
  Recognition}{on~Document~Analysis and Recognition}{2019}]%
        {icdar}
\bibfield{author}{\bibinfo{person}{International~Conference on
  Document~Analysis} {and} \bibinfo{person}{Recognition}.}
  \bibinfo{year}{2019}\natexlab{}.
\newblock \bibinfo{booktitle}{\emph{SROIE, ICADR,
  \url{https://rrc.cvc.uab.es/?ch=13&com=introduction}, 2019}}.
\newblock


\bibitem[\protect\citeauthoryear{Palm, Laws, and Winther}{Palm
  et~al\mbox{.}}{2018}]%
        {palm2018attend}
\bibfield{author}{\bibinfo{person}{Rasmus~Berg Palm}, \bibinfo{person}{Florian
  Laws}, {and} \bibinfo{person}{Ole Winther}.} \bibinfo{year}{2018}\natexlab{}.
\newblock \showarticletitle{Attend, Copy, Parse-End-to-end information
  extraction from documents}.
\newblock \bibinfo{journal}{\emph{arXiv preprint arXiv:1812.07248}}
  (\bibinfo{year}{2018}).
\newblock


\bibitem[\protect\citeauthoryear{Peanho, Stagni, et~al\mbox{.}}{Peanho
  et~al\mbox{.}}{[n.d.]}]%
        {peanho2012semantic}
\bibfield{author}{\bibinfo{person}{Claudio~Antonio Peanho},
  \bibinfo{person}{Henrique Stagni}, {et~al\mbox{.}}}
  \bibinfo{year}{[n.d.]}\natexlab{}.
\newblock \showarticletitle{Semantic information extraction from images of
  complex documents}.
\newblock \bibinfo{journal}{\emph{Applied Intelligence}} \bibinfo{volume}{37},
  \bibinfo{number}{4} (\bibinfo{year}{[n.\,d.]}).
\newblock


\bibitem[\protect\citeauthoryear{Ravi and Larochelle}{Ravi and
  Larochelle}{2017}]%
        {ravi2016optimization}
\bibfield{author}{\bibinfo{person}{Sachin Ravi} {and} \bibinfo{person}{Hugo
  Larochelle}.} \bibinfo{year}{2017}\natexlab{}.
\newblock \showarticletitle{Optimization as a model for few-shot learning}.
\newblock \bibinfo{journal}{\emph{International Conference on Learning
  Representations}} (\bibinfo{year}{2017}).
\newblock


\bibitem[\protect\citeauthoryear{Rusinol, Benkhelfallah, et~al\mbox{.}}{Rusinol
  et~al\mbox{.}}{2013}]%
        {rusinol2013field}
\bibfield{author}{\bibinfo{person}{Mar{\c{c}}al Rusinol},
  \bibinfo{person}{Tayeb Benkhelfallah}, {et~al\mbox{.}}}
  \bibinfo{year}{2013}\natexlab{}.
\newblock \showarticletitle{Field extraction from administrative documents by
  incremental structural templates}. In \bibinfo{booktitle}{\emph{International
  Conference on Document Analysis and Recognition}}.
\newblock


\bibitem[\protect\citeauthoryear{Salton and Buckley}{Salton and
  Buckley}{1988}]%
        {salton1988term}
\bibfield{author}{\bibinfo{person}{Gerard Salton} {and}
  \bibinfo{person}{Christopher Buckley}.} \bibinfo{year}{1988}\natexlab{}.
\newblock \showarticletitle{Term-weighting approaches in automatic text
  retrieval}.
\newblock \bibinfo{journal}{\emph{Information processing \& management}}
  \bibinfo{volume}{24}, \bibinfo{number}{5} (\bibinfo{year}{1988}),
  \bibinfo{pages}{513--523}.
\newblock


\bibitem[\protect\citeauthoryear{Santoro, Bartunov, et~al\mbox{.}}{Santoro
  et~al\mbox{.}}{2016}]%
        {santoro2016meta}
\bibfield{author}{\bibinfo{person}{Adam Santoro}, \bibinfo{person}{Sergey
  Bartunov}, {et~al\mbox{.}}} \bibinfo{year}{2016}\natexlab{}.
\newblock \showarticletitle{Meta-learning with memory-augmented neural
  networks}. In \bibinfo{booktitle}{\emph{International Conference on Machine
  Learning}}.
\newblock


\bibitem[\protect\citeauthoryear{Shi, Bai, and Yao}{Shi et~al\mbox{.}}{2016}]%
        {shi2016end}
\bibfield{author}{\bibinfo{person}{Baoguang Shi}, \bibinfo{person}{Xiang Bai},
  {and} \bibinfo{person}{Cong Yao}.} \bibinfo{year}{2016}\natexlab{}.
\newblock \showarticletitle{An end-to-end trainable neural network for
  image-based sequence recognition and its application to scene text
  recognition}.
\newblock \bibinfo{journal}{\emph{IEEE Transactions on Pattern Analysis and
  Machine Intelligence}} (\bibinfo{year}{2016}).
\newblock


\bibitem[\protect\citeauthoryear{Snell, Swersky, and Zemel}{Snell
  et~al\mbox{.}}{2017}]%
        {snell2017prototypical}
\bibfield{author}{\bibinfo{person}{Jake Snell}, \bibinfo{person}{Kevin
  Swersky}, {and} \bibinfo{person}{Richard Zemel}.}
  \bibinfo{year}{2017}\natexlab{}.
\newblock \showarticletitle{Prototypical networks for few-shot learning}. In
  \bibinfo{booktitle}{\emph{Neural Information Processing Systems}}.
\newblock


\bibitem[\protect\citeauthoryear{Soto and Yoo}{Soto and Yoo}{2019}]%
        {soto2019visual}
\bibfield{author}{\bibinfo{person}{Carlos Soto} {and} \bibinfo{person}{Shinjae
  Yoo}.} \bibinfo{year}{2019}\natexlab{}.
\newblock \showarticletitle{Visual Detection with Context for Document Layout
  Analysis}. In \bibinfo{booktitle}{\emph{Proceedings of the 2019 Conference on
  Empirical Methods in Natural Language Processing and the 9th International
  Joint Conference on Natural Language Processing (EMNLP-IJCNLP)}}.
  \bibinfo{pages}{3455--3461}.
\newblock


\bibitem[\protect\citeauthoryear{Sunder, Srinivasan, et~al\mbox{.}}{Sunder
  et~al\mbox{.}}{2019}]%
        {sunder2019one}
\bibfield{author}{\bibinfo{person}{Vishal Sunder}, \bibinfo{person}{Ashwin
  Srinivasan}, {et~al\mbox{.}}} \bibinfo{year}{2019}\natexlab{}.
\newblock \showarticletitle{One-shot Information Extraction from Document
  Images using Neuro-Deductive Program Synthesis}.
\newblock \bibinfo{journal}{\emph{International Joint Conference on Artificial
  Intelligence workshops}} (\bibinfo{year}{2019}).
\newblock


\bibitem[\protect\citeauthoryear{Sung, Yang, et~al\mbox{.}}{Sung
  et~al\mbox{.}}{2018}]%
        {sung2018learning}
\bibfield{author}{\bibinfo{person}{Flood Sung}, \bibinfo{person}{Yongxin Yang},
  {et~al\mbox{.}}} \bibinfo{year}{2018}\natexlab{}.
\newblock \showarticletitle{Learning to compare: Relation network for few-shot
  learning}. In \bibinfo{booktitle}{\emph{IEEE Conference on Computer Vision
  and Pattern Recognition}}.
\newblock


\bibitem[\protect\citeauthoryear{Veli{\v{c}}kovi{\'c}, Cucurull, Casanova,
  Romero, Lio, and Bengio}{Veli{\v{c}}kovi{\'c} et~al\mbox{.}}{2018}]%
        {velivckovic2017graph}
\bibfield{author}{\bibinfo{person}{Petar Veli{\v{c}}kovi{\'c}},
  \bibinfo{person}{Guillem Cucurull}, \bibinfo{person}{Arantxa Casanova},
  \bibinfo{person}{Adriana Romero}, \bibinfo{person}{Pietro Lio}, {and}
  \bibinfo{person}{Yoshua Bengio}.} \bibinfo{year}{2018}\natexlab{}.
\newblock \showarticletitle{Graph attention networks}.
\newblock \bibinfo{journal}{\emph{International Conference on Learning
  Representations}} (\bibinfo{year}{2018}).
\newblock


\bibitem[\protect\citeauthoryear{Vinyals, Bengio, and Kudlur}{Vinyals
  et~al\mbox{.}}{2015}]%
        {vinyals2015order}
\bibfield{author}{\bibinfo{person}{Oriol Vinyals}, \bibinfo{person}{Samy
  Bengio}, {and} \bibinfo{person}{Manjunath Kudlur}.}
  \bibinfo{year}{2015}\natexlab{}.
\newblock \showarticletitle{Order matters: Sequence to sequence for sets}.
\newblock \bibinfo{journal}{\emph{International Conference on Learning
  Representations}} (\bibinfo{year}{2015}).
\newblock


\bibitem[\protect\citeauthoryear{Vinyals, Blundell, et~al\mbox{.}}{Vinyals
  et~al\mbox{.}}{2016}]%
        {vinyals2016matching}
\bibfield{author}{\bibinfo{person}{Oriol Vinyals}, \bibinfo{person}{Charles
  Blundell}, {et~al\mbox{.}}} \bibinfo{year}{2016}\natexlab{}.
\newblock \showarticletitle{Matching networks for one shot learning}. In
  \bibinfo{booktitle}{\emph{Neural Information Processing Systems}}.
\newblock


\bibitem[\protect\citeauthoryear{Xu, Li, Cui, Huang, Wei, and Zhou}{Xu
  et~al\mbox{.}}{2019}]%
        {xu2019layoutlm}
\bibfield{author}{\bibinfo{person}{Yiheng Xu}, \bibinfo{person}{Minghao Li},
  \bibinfo{person}{Lei Cui}, \bibinfo{person}{Shaohan Huang},
  \bibinfo{person}{Furu Wei}, {and} \bibinfo{person}{Ming Zhou}.}
  \bibinfo{year}{2019}\natexlab{}.
\newblock \bibinfo{title}{LayoutLM: Pre-training of Text and Layout for
  Document Image Understanding}.
\newblock
\newblock
\showeprint[arxiv]{1912.13318}~[cs.CL]


\bibitem[\protect\citeauthoryear{Yang, Cheng, et~al\mbox{.}}{Yang
  et~al\mbox{.}}{2018}]%
        {yang2018inceptext}
\bibfield{author}{\bibinfo{person}{Qiangpeng Yang}, \bibinfo{person}{Mengli
  Cheng}, {et~al\mbox{.}}} \bibinfo{year}{2018}\natexlab{}.
\newblock \showarticletitle{Inceptext: A new inception-text module with
  deformable psroi pooling for multi-oriented scene text detection}.
\newblock \bibinfo{journal}{\emph{International Joint Conference on Artificial
  Intelligence}} (\bibinfo{year}{2018}).
\newblock


\bibitem[\protect\citeauthoryear{Yang, Yumer, et~al\mbox{.}}{Yang
  et~al\mbox{.}}{2017}]%
        {yang2017learning}
\bibfield{author}{\bibinfo{person}{Xiao Yang}, \bibinfo{person}{Ersin Yumer},
  {et~al\mbox{.}}} \bibinfo{year}{2017}\natexlab{}.
\newblock \showarticletitle{Learning to extract semantic structure from
  documents using multimodal fully convolutional neural networks}. In
  \bibinfo{booktitle}{\emph{IEEE Conference on Computer Vision and Pattern
  Recognition}}.
\newblock


\bibitem[\protect\citeauthoryear{Zanfir and Sminchisescu}{Zanfir and
  Sminchisescu}{2018}]%
        {zanfir2018deep}
\bibfield{author}{\bibinfo{person}{Andrei Zanfir} {and}
  \bibinfo{person}{Cristian Sminchisescu}.} \bibinfo{year}{2018}\natexlab{}.
\newblock \showarticletitle{Deep learning of graph matching}. In
  \bibinfo{booktitle}{\emph{Proceedings of the IEEE Conference on Computer
  Vision and Pattern Recognition}}. \bibinfo{pages}{2684--2693}.
\newblock


\bibitem[\protect\citeauthoryear{Zhou, Yao, et~al\mbox{.}}{Zhou
  et~al\mbox{.}}{2017}]%
        {zhou2017east}
\bibfield{author}{\bibinfo{person}{Xinyu Zhou}, \bibinfo{person}{Cong Yao},
  {et~al\mbox{.}}} \bibinfo{year}{2017}\natexlab{}.
\newblock \showarticletitle{EAST: an efficient and accurate scene text
  detector}. In \bibinfo{booktitle}{\emph{IEEE Conference on Computer Vision
  and Pattern Recognition}}. \bibinfo{pages}{5551--5560}.
\newblock


\end{thebibliography}
